\documentclass[authoryear,preprint,review,12pt]{elsarticle}
\usepackage{hyperref}
\usepackage{url}
\usepackage{booktabs}       
\usepackage{amsfonts}       
\usepackage{nicefrac}       
\usepackage{microtype}      
\usepackage{graphicx}
\usepackage{amsmath}
\usepackage{multirow}
\usepackage[utf8]{inputenc} 
\usepackage[T1]{fontenc}    
\usepackage{color}
\usepackage{algorithm}
\usepackage{algorithmic}
\usepackage{amssymb}

\journal{Pattern Recognition}

\usepackage{natbib}
\setcitestyle{numbers,square}

\begin{document}

\begin{frontmatter}

\title{FG-UAP: Feature-Gathering Universal Adversarial Perturbation}


\author[mymainaddress]{Zhixing Ye}
\ead{yzx1213@sjtu.edu.cn}

\author[mymainaddress]{Xinwen Cheng}
\ead{xinwencheng@sjtu.edu.cn}

\author[mymainaddress]{Xiaolin Huang\corref{mycorrespondingauthor}}
\cortext[mycorrespondingauthor]{Corresponding author}
\ead{xiaolinhuang@sjtu.edu.cn}

\address[mymainaddress]{Department of Automation 
and the MOE Key Laboratory of System Control and Information Processing, Shanghai Jiao Tong University, Shanghai, 200240, P.R. China.}

\begin{abstract}

Deep Neural Networks (DNNs) are susceptible to elaborately designed perturbations, whether such perturbations are dependent or independent of images. The latter one, called Universal Adversarial Perturbation (UAP), is very attractive for model robustness analysis, since its independence of input reveals the intrinsic characteristics of the model. Relatively, another interesting observation is Neural Collapse (NC), which means the feature variability may collapse during the terminal phase of training. Motivated by this, we propose to generate UAP by attacking the layer where NC phenomenon happens. Because of NC, the proposed attack could gather all the natural images' features to its surrounding, which is hence called Feature-Gathering UAP (FG-UAP).

We evaluate the effectiveness our proposed algorithm on abundant experiments, including untargeted and targeted universal attacks, attacks under limited dataset, and transfer-based black-box attacks among different architectures including Vision Transformers, which are believed to be more robust. Furthermore, we investigate FG-UAP in the view of NC by analyzing the labels and extracted features of adversarial examples, finding that collapse phenomenon becomes stronger after the model is corrupted. \textit{The code will be released when the paper is accepted.}

\end{abstract}


\end{frontmatter}

\section{Introduction}

Deep Neural Networks (DNNs) have been proved to be vulnerable to adversarial perturbations \cite{bfgs, fgsm, pgd, deepfool, cw, onepixel, hang2020ensemble, huang2022cyclical}. In a classification task, for a sample correctly predicted by a DNN, it can be easily predicted as a wrong class after adding an imperceptible perturbation elaborately crafted by the attacker. However, such kind of perturbations are image dependent, which means the attackers 
craft perturbations corresponding to each datum. Later, Moosavi-Dezfooli et al. \cite{deepfooluap}, for the first time, proposed a special type of adversarial attack,
which 
is to fool the DNNs by adding the same perturbation, called Universal Adversarial Perturbation (UAP), to all samples. 
Since its proposal,
researchers have figured out numerous ways to craft UAPs \cite{nag, gap, gduap, aaa, pdua, uapforod}. 

Compared with regular adversarial examples, which reveal the over-sensitivity of DNNs, the UAPs are different: they reveal that DNNs could be largely affected by a single perturbation for almost all input. The mechanism relies on the guess that the difference for different images from even different classes is vanishing, such that they can be attacked by the same perturbation. This guess coincides with the recently discovered \emph{Neural Collapse} (NC, \cite{neuralcollapse}), which means that 
a DNN induces an underlying mathematical simplicity to the last-layer activation. 
One of NC manifestations is variability collapse, where the within-class variation of the activations becomes negligible as these activations collapse to their class means, from which it follows that finding an UAP for all the samples may be more feasible.  

The collapse of the difference among samples is the essential reason why we can find universal perturbations. Thus, directly attacking the layers where NC happens is expected to have stronger UAPs than attacking other places, e.g., the output like the most UAP methods. This is just what we want to do in this paper. Specifically,  
with a proposed Feature-Gathering loss (FG loss), the adversary manages to find stronger universal perturbations in the layer which owns little within-class diversity and meanwhile expressive for potential perturbations. Our method is simple but effective and is verified by numerical experiments that we can outperform  
the state-of-the-art UAPs, 
whether untargeted or targeted attackes, in both regular and  
limited training datasets. 
Moreover, with the proposed method, we can 
generate UAPs for Vision Transformers (ViTs) \cite{vit}, which are free from convolutional architectures and are believed to be more robust against adversarial perturbations \cite{mahmood2021robustness, shao2021adversarial, aldahdooh2021reveal}. Results show that our method can also defeat cutting-edge baselines, though ViTs are indeed less likely to be fooled by UAPs compared with convolutional neural networks.  We further evaluate the transferability among CNNs and ViTs, discovering that CNNs can be more easily attacked by UAPs calculated for other structures including ViTs, while not vise versa.

Not only for generating stronger attacks, we can also use the proposed UAPs to better investigate the DNNs. By analyzing the features of UAPs, we find a new collapse phenomenon that features of UAPs concentrate to a direction in the layer we exert attacks. This provides a new evidence for NC and can explain the phenomenon of dominant labels, which is mentioned but not fully discussed in \cite{deepfooluap, cosineuap}. 


\subsection*{Contributions}
\begin{itemize}
  \item 
  Inspired by the NC phenomenon, we propose a simple but effective method to generate strong UAPs for DNNs. 
  We name this UAP as Feature-Gathering UAP (FG-UAP) for its strong ability to gather natural images' features.
  \item We verify the effectiveness of our FG-UAP on various DNNs and achieve state-of-the-art performance not only in untargeted task but also in targeted and mini-set tasks. 
  \item We discuss the mechanism of FG-UAP in the view of NC by analyzing the labels and features of adversarial examples extracted by DNNs, providing a more detailed explanation on the dominant label phenomenon. 
\end{itemize}

\section{Related Work}\label{related}

Szegedy \textit{et al. }\cite{bfgs} firstly observed that DNNs are vulnerable to maliciously constructed small noises called adversarial perturbations. Following this discovery, numerous attack methods have been proposed, including Fast Gradient Sign Method (FGSM) \cite{fgsm}, Projected Gradient Descent (PGD) \cite{pgd}, C\&W \cite{cw}, DeepFool \cite{deepfool}, and AoA \cite{aoa}. These methods craft perturbations by designing different losses and optimization algorithms, and have been extended to various research fields \cite{ma2021understanding, bisogni2021adversarial, li2021black}.
Notice that perturbations generated by all the above methods are image-dependent, which means different perturbations must be specifically computed for different images. 

Different from image-dependent attacks, image-agnostic attacks shift the majority of images' predictions with a single perturbation, named Universal Adversarial Perturbation (UAP). This special type of adversarial perturbations is firstly proposed by Moosavi-Dezfooli\textit{ et al.} \cite{deepfooluap}, where an iterative procedure based on Deepfool \cite{deepfool} is designed. To distinguish this type from other UAPs, we hereinafter refer to this UAP method as DeepFool-UAP. Motivated by \cite{deepfooluap}, researchers proposed more algorithms to generate UAPs. Mopuri \textit{et al.} \cite{nag} put forward a Network for Adversary Generation (NAG) to model the distribution of adversarial perturbations. Omid Poursaeed \textit{et al.} \cite{gap} present Generative Adversarial Perturbations (GAP) to create UAPs for both classification and semantic segmentation tasks. In addition, they are the first to present challenging targeted UAPs. Later, it has been confirmed that targeted UAPs can also be found by exploiting a proxy dataset instead of the original training data \cite{dfuap}. Mopuri \textit{et al.} \cite{gduap} compute UAPs by overfiring the extracted features at multiple layers. Liu \textit{et al.} \cite{pdua} consider the model uncertainty to craft an insensitive universal perturbation. Li \textit{et al.} \cite{uapforod} try to extend such universal attack to detector-level. The latest Cosine-UAP \cite{cosineuap} proposes an algorithm based on cosine similarity to craft the state-of-the-art UAP, and also discusses the phenomenon of dominant class, which is firstly discovered by \cite{deepfooluap}.

Neural Collapse (NC) becomes another line of research, since Papyan \textit{et al. }\cite{neuralcollapse} first revealed the tendency to a simple symmetric structure in penultimate layers during the terminal phase of training. The empirical demystifying of penultimate features has spurred extensive research on theoretical philosophy underlying in different settings. A literature of study admirably proves that global minimizers of cross-entropy \cite{weinan2022emergence,lu2022neural}, MSE \cite{han2022neural}, and contrastive loss \cite{awasthi2022more} are all NC favorable. Zhu \textit{et al.} \cite{zhu2021geometric} elucidate that any optimization algorithm which can escape strict saddle points will converge to NC, showcasing SGD \cite{saad1998online},  Adam \cite{kingma2014adam} and LBFGS \cite{bottou2018optimization} through experimental verification. \cite{tirer2022extended} generalizes previous results by adding a nonlinear layer, presenting the same succinct structure as before. 
This venerable line of work corroborates that NC persists across a wide range of well-trained overparameterized neural networks. It is instructive to associate practical implications of NC with adversarial attacks. Thus, we make early attempts on utilizing this pervasive behavioral simplicity of high-level features to generate UAPs for the first time. 

\section{Proposed Approach}

\subsection{Problem formulation}


Consider a classification task with 
natural images $X = \{\boldsymbol{x_1}, \boldsymbol{x_2}, \dots, \boldsymbol{x_n}\}$ and the corresponding labels $y_1, y_2, \dots, y_n$. A classifier $C(\cdot)$ maps an input image $x_i$ to an estimated label $y_i=C(\boldsymbol{x_i})$. The goal of the universal attack is to fool the classifier with a single perturbation. This means the 
victim classifier prones to predict any image as an incorrect 
class when this image is corrupted 
by this perturbation. Mathematically,  finding such UAP  
$\boldsymbol{\delta}$ can be formulated as the following problem, 
\begin{eqnarray}
    \begin{array}{ll}
     &  \mathop{\text{max}}\limits_{\boldsymbol{\delta}} ~ \mathbb{P}_{\boldsymbol{x} \in X}[C(\boldsymbol{x}) \neq C(\boldsymbol{x} + \boldsymbol{\delta})]\\
         & \text{s.t.} ~ \|\boldsymbol{\delta} \|\leq \xi, 
    \end{array}
\end{eqnarray}
where $\xi$ is a user-given threshold 
to ensure that the perturbation is visually imperceptible to humans.

For a typical classifier, it can be composed of an feature extractor with numerous layers, an average pooling layer, followed by one or more fully connected linear layers, as shown in Figure \ref{fig:framework}. Except for the last linear layer, all the linear layers are followed by non-linear activations. 
Denote the layers before the last linear layer as calculating a function $\boldsymbol{h}$. This function maps an input $\boldsymbol{x}$ to a feature vector $\boldsymbol{h(x)}$, which is referred to as the \textit{last-layer feature} according to \cite{neuralcollapse}. The final output of the last layer contains the logit value for each class, which is usually called the logit vector, and this output space is called logit space. Logit space is the output of an end-to-end DNN, numerous attack methods are implemented based on this space \cite{deepfooluap, rhp, aaa, gap, fastuap, dfuap, cosineuap}.

\subsection{Neural Collapse and UAP}
Recently, Neural Collapse (NC, \cite{neuralcollapse}) has been found as a 
special and essential phenomenon happened in the last-layer activation in DNNs. A manifestation of NC is the variability collapse, i.e., the within-class variation of the activations becomes negligible as these activations collapse to their class means: 
\begin{equation}
\boldsymbol{\Sigma}_{W} \triangleq \underset{i, c}{\operatorname{Ave}}\left\{\left(\boldsymbol{h}_{i, c}-\boldsymbol{\mu}_{c}\right)\left(\boldsymbol{h}_{i, c}-\boldsymbol{\mu}_{c}\right)^{\top}\right\}\rightarrow 0,
\end{equation}
where $\boldsymbol{h}_{i,c}\triangleq\boldsymbol{h}(\boldsymbol{x}_{i,c})$ is the last-layer feature of the $i$-th sample in the $c$-th class and $\boldsymbol{\mu}_{c}\triangleq \underset{i}{\operatorname{Ave}}\{\boldsymbol{h}_{i,c}\}$ is the mean value of the corresponding class.  
Such collapse of the samples' difference 
perfectly explains the existence of UAPs: The diversity of natural images is dampen in a well-trained DNN, from which it follows that 
one can find 
perturbations that exist in a 
subspace in which most of the normal vectors of decision boundaries lie. 
Such a perturbation 
can then fool the majority of other images with the same class. 

As long as there is NC, we can find the corresponding UAPs more easily. An extreme case is that NC happens in the output layer, where images in the same class naturally have no variance, which results in the current UAPs. however, from the view of attack, we prefer a more expressive space to attack so that we have more freedom to choose a good attack direction. Now that NC could happen not only in the output layer but also in the last-layer feature space,
we deem that attacking at the last-layer feature space is a more effective and significative choice. 

\subsection{Feature-Gathering UAP (FG-UAP)}

\begin{figure}[htbp]
    \centering
    \includegraphics[width=.9\textwidth]{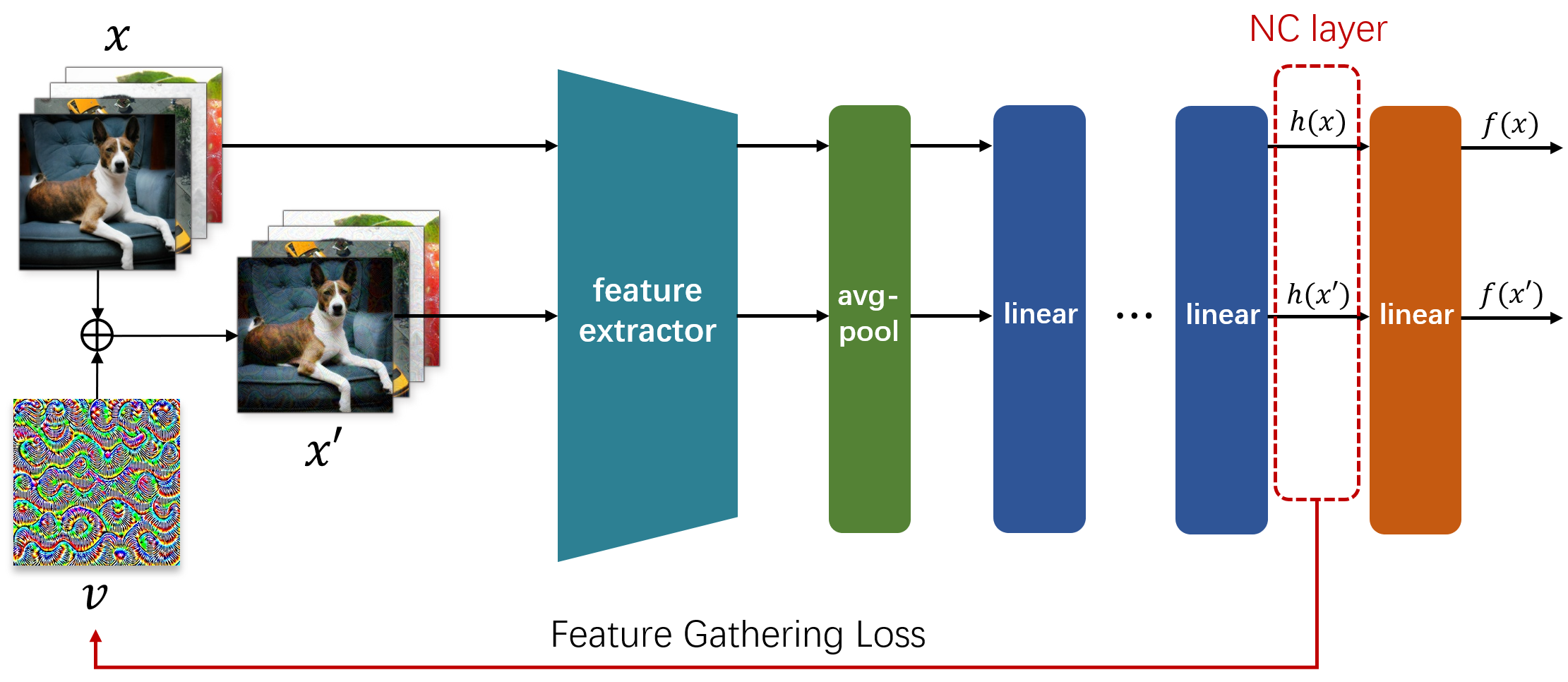}
    \caption{The framework of FG-UAP. For a batch of training images $\boldsymbol{x}$, we first input them into the DNN to get their last-layer features. Then, add the universal perturbation $\boldsymbol{\delta}$ to the images to get a set of adversarial examples $\boldsymbol{x'}$ and get their last-layer features. Finally, calculate FG-Loss with these features and optimize the $\boldsymbol{\delta}$ iteratively. }
    \label{fig:framework}
\end{figure}
Now we explain the details of how to craft UAPs by attacking the last-layer feature space. Practically, we take turns to input clean samples into the targeted DNN, and lower the similarity of $\boldsymbol{h}(\boldsymbol{x})$ and $\boldsymbol{h}(\boldsymbol{x}+\boldsymbol{\delta})$. In terms of measuring similarity, we use cosine similarity, considering its adaption to vector scene and effectiveness proved in previous researches \cite{cosineuap, dp}.  To this end, we design a loss named Feature-Gathering loss (FG loss). Given a natural sample $\boldsymbol{x}$ and its corresponding adversarial example $\boldsymbol{x} + \boldsymbol{\delta}$, their features at the last-layer feature space are abbreviated as $\boldsymbol{h}$ and $\boldsymbol{h'}$, respectively. Without any other label information, we can directly compute the FG loss:
\begin{equation}
    \mathcal{L}_{\mathrm{FG}}(\boldsymbol{h}, \boldsymbol{h'}) = \text{sim}(\boldsymbol{h}, \boldsymbol{h'}) = \frac{\boldsymbol{h} \cdot \boldsymbol{h'}}{\|\boldsymbol{h}\|\| \boldsymbol{h'}\|}
    = \frac{\sum\limits_{i=1}^{n}h_i {h_i'}}
    {\sqrt{\sum\limits_{i=1}^{n}h_i^2}\sqrt{\sum\limits_{i=1}^{n}{h_i'}^2}}
\end{equation}
The whole attacking procedure is demonstrated in Figure \ref{fig:framework} and the algorithm is detailed in Algorithm \ref{ag:fguap}. Since the crafted UAP has the ability to gather natural images' features to a new direction (refer to \ref{sec:property} for experiment details), we name this universal perturbation as Feature-Gathering UAP (FG-UAP).
\begin{algorithm}[htbp]
\caption{FG-UAP}
\label{ag:fguap}
\textbf{Input}: classifier $f$, training set $X$, perturbation magnitude $\xi$, batch size $b$, maximum number of epochs $m$, learning rate optimizer $lr$.\\
\textbf{Output}: universal perturbation $\boldsymbol{\delta}$ 
\begin{algorithmic}[1]
\STATE Initialize $\boldsymbol{\delta}\leftarrow 0, t \leftarrow 0$.
\WHILE{$t<m$}
    \FOR{each batch of data $B_i \subset X:len(B_i)=b$}
        \STATE $\boldsymbol{g} \leftarrow \underset{\boldsymbol{x}\sim B_i}{\mathbb{E}}\left[\nabla_{\boldsymbol{\delta}}\mathcal{L}_{\mathrm{FG}}(\boldsymbol{h}(\boldsymbol{x}), \boldsymbol{h}(\boldsymbol{x+\delta}))\right]$
        \STATE $\boldsymbol{\delta} \leftarrow \boldsymbol{\delta} + \Gamma (\boldsymbol{g},lr)$  ~~~~\text{\# update the perturbation with the optimizer}
        \STATE $\boldsymbol{\delta} \leftarrow \text{clamp}(\boldsymbol{\delta},  -\xi, \xi)$ ~~~~\text{\# clamp the perturbation}
    \ENDFOR
    \STATE $t \leftarrow t+1$
\ENDWHILE
\end{algorithmic}
\end{algorithm}

\section{Experiments}\label{expsec}

In numerical experiments, we mainly evaluate the proposed FG-UAP 
for six typical DNNs with convolutional architectures (hereinafter abbreviated as CNNs). They are pre-trained on the ILSVRC 2012 \cite{imagenet} validation set ($50,000$ images). The 
victim CNNs include AlexNet \cite{alexnet}, GoogLeNet \cite{googlenet}, VGG16 \cite{vgg}, VGG19 \cite{vgg}, ResNet50 \cite{resnet}, and ResNet152 \cite{resnet}, all of which
are got from Torchvision \cite{pytorch}. 
Adam \cite{kingma2014adam} is chosen as the optimizer. The hyper-parameters in Algorithm \ref{ag:fguap} are set as $b=32$, $m=10$, $lr=0.02$, and the magnitude of crafted UAP is set as $\xi = 10/255$, which is consistent with other universal attack methods. All the experiments are performed on PyTorch \cite{pytorch} with NVIDIA GeForce RTX 2080Ti GPUs. 

\subsection{Universal attack on CNNs} \label{sec:cnn}
\begin{figure}[H]
    \centering
    \includegraphics[width=\textwidth]{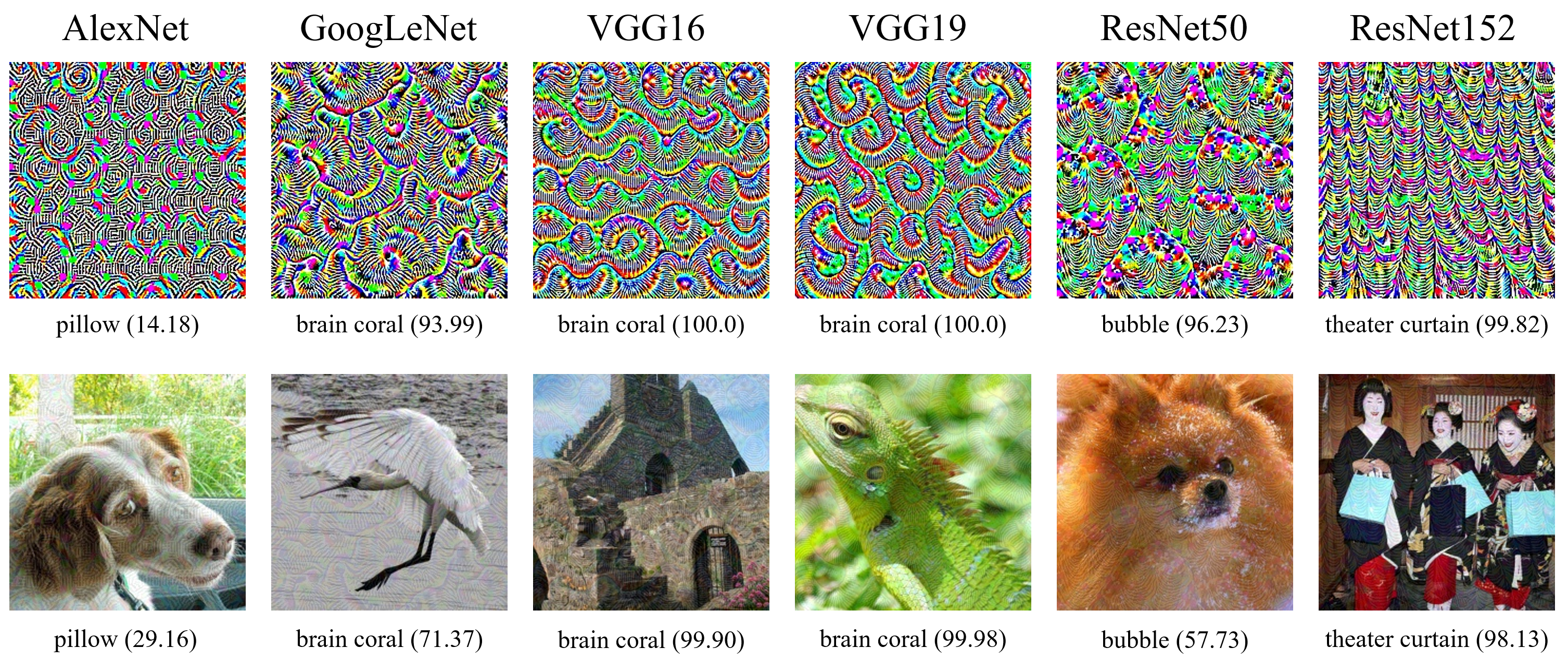}
    \vspace{-5mm}
    \caption{FG-UAPs (Magnitudes are mapped from $[-10,10]$ to $[0,255]$ for better observation.) and 
    corresponding perturbed images for different CNNs. Prediction and confidence (\%) are also reported under each image for reference. }
    \label{fig:cnn_attack}
\end{figure}

We first train FG-UAP to attack CNNs 
and calculate their Fooling Ratios (FRs), 
the primary criterion to evaluate the strength of a UAP.  Table \ref{tab:cnn} shows the experimental results along with other UAP methods. From the table, it can be observed that FG-UAP achieves the highest FRs for all victim models and the improvements from state-of-the-art methods are at least 1\% and up to 5\%.
Since the logit function and the optimization method of FG-UAP are not unique, actually have appeared in different methods, the good performance 
verifies our expectation that attacking the layers with variability collapse results in stronger universal perturbations. Figure \ref{fig:cnn_attack} visualizes the generated FG-UAPs and corresponding sample adversarial images for different CNNs. 
\begin{table}[H]
\centering
\caption{FRs (\%) of different UAP methods attacking CNNs on ImageNet validation set, where the best FRs are highlighted in bold, and the second-best ones are underlined.}
\label{tab:cnn}
\vspace{2mm}
\resizebox{\linewidth}{!}{
\begin{tabular}{@{}ccccccc@{}}
\toprule
Method     & AlexNet            & GoogLeNet             & VGG16          & VGG19          & ResNet50           & ResNet152          \\ \midrule
DeepFool-UAP \cite{deepfooluap}        & 93.3           & 78.9           & 78.3           & 77.8           & -              & 84.0           \\
GAP \cite{gap}        & -              & 82.7           & 83.7           & 80.1           & 62.8              & 59.19              \\
NAG \cite{nag}        & 96.44              & 90.37          & 77.57          & 83.78          & 86.64          & 87.24          \\
FTUAP \cite{ftuap}      & -              & 85.8           & 93.5           & 94.5           & 93.6           & \underline{92.7}           \\
DF-UAP \cite{dfuap}     & 96.17          & 88.94          & 94.30          & 94.98          & \underline{94.96}              & 90.08          \\
Cosine-UAP \cite{cosineuap} & \underline{96.5}           & \underline{90.5}           & \underline{97.4}           & \underline{96.4}           & -              & 90.2           \\
FG-UAP     & \textbf{97.77} & \textbf{91.53} & \textbf{98.45} & \textbf{97.77} & \textbf{96.23} & \textbf{95.59} \\ \bottomrule
\end{tabular}}
\end{table}

\subsection{Universal attack on ViTs}
 Unlike CNNs, ViTs \cite{vit} use 
 pure self-attention-based architectures instead of convolutional blocks, 
 which is now believed 
 to  
 enhance robustness. Thus, it is of great necessity to confirm whether such structure is vulnerable to our attack method. In this experiment, we apply  Algorithm \ref{ag:fguap} (with parameters $b=32$, $m=20$, $lr=0.01$ and the rest are remain unchanged) on DeiT family \cite{deit}. Since DeiT-Ti and DeiT-S can be regarded as the counterpart of ResNet50 and ResNet18, respectively, we also consider these two models.
 For comparison, we list the FRs for Cosine-UAP, the state-of-the-art UAP method. 
 Experimental results are shown in Table \ref{tab:vit}, where we have two observations: i) ViTs are indeed more robust than counterpart CNNs; ii) FG-UAP can still get satisfactory FRs and has more significant advantages on attacking ViTs, for which Cosine-UAP's performance largely degrades, especially for DeiT-B.
\begin{table}[!htpb]
\centering
\caption{FRs (\%) of FG-UAP and state-of-the-art method attacking ViTs on the ImageNet validation set. Results for ResNet18 and ResNet50 are reported here since their scales are comparable to DeiT-Ti and DeiT-S, respectively. The best FRs are highlighted in bold.}
\label{tab:vit}
\vspace{2mm}
\begin{tabular}{@{}c|ccc|ccc@{}}
\toprule
        & DeiT-Ti & DeiT-S & DeiT-B & ResNet18 & ResNet50 \\ \midrule
Cosine-UAP  & 90.75   & 81.13    & 69.25  & 94.72   & 95.44 \\
FG-UAP &  \textbf{92.54}   & \textbf{83.47}     & \textbf{85.58} &\textbf{95.37} & \textbf{96.43}      \\ \bottomrule
\end{tabular}
\end{table}

\subsection{Cross-model transferability of FG-UAP}

Cross-model transferability is another criterion for UAPs. To evaluate this point, we train FG-UAPs for one architecture and use others to classify them. The results are displayed in Table \ref{tab:transfer}, where  ALN, GLN, RN stands for AlexNet, GoogLeNet, and ResNet, respectively. The result clearly shows that UAPs trained for other CNNs keep quite high FRs for CNNs. Even for ViTs with totally different structures, the generated UAPs have strong transferability to CNNs, especially AlexNet and VGGs. On the contrary, transfer-based black-box attacks for ViTs are much harder: the best FR is only 36.79\%, even when the transfer is among similar architectures. These results implies that DNNs with convolutional blocks may share similar vulnerability so that the universal attack are more easily to be transferred, while that is not the case for self-attention-based architectures.



\begin{table}[H]
\centering
\caption{Transferability of FG-UAP across different DNNs. The rows indicate the surrogate model to compute UAPs, and the columns indicate the victim models for which FRs (\%) are reported.}
\label{tab:transfer}
\vspace{2mm}
\resizebox{\linewidth}{!}{
\begin{tabular}{cccccccccc} 
\toprule
       & ALN   & GLN   & VGG16 & VGG19 & RN50  & RN152 & DeiT-Ti & DeiT-S & DeiT-B  \\
\midrule       
ALN    & \textbf{97.77} & 55.77 & 69.40 & 63.90 & 48.76 & 39.47 & 29.85  & 19.02  & 15.13   \\
GLN    & 53.20 & \textbf{91.52} & 76.06 & 73.14 & 59.64 & 49.02 & 32.26  & 22.03  & 14.80   \\
VGG16  & 46.50 & 53.04 & \textbf{98.44} & 93.46 & 56.74 & 45.76 & 25.43  & 18.80  & 13.83   \\
VGG19  & 48.17 & 54.64 & 95.53 & \textbf{97.77} & 59.70 & 49.34 & 27.36  & 20.15  & 13.63   \\
RN50   & 52.57 & 59.69 & 79.15 & 75.77 & \textbf{96.23} & 65.16 & 25.64  & 16.81  & 12.65   \\
RN152  & 51.30 & 65.42 & 86.05 & 83.05 & 84.61 & \textbf{95.48} & 27.46  & 18.54  & 14.44   \\
DeiT-Ti & 43.26 & 28.53 & 49.59 & 46.96 & 31.11 & 22.47 & \textbf{92.54}  & 24.49  & 20.48   \\
DeiT-S & 50.65 & 39.89 & 53.35 & 51.64 & 35.96 & 29.92 & 36.23  & \textbf{83.47}  & 21.41   \\
DeiT-B & 53.06 & 43.12 & 56.72 & 55.11 & 37.42 & 32.70 & 36.79  & 32.84  & \textbf{85.58}   \\
\bottomrule
\end{tabular}}
\end{table}
 
 \subsection{Mini-set UAP and Targeted UAP}\label{sec:mini}
 Here we also test FG-UAP on two variant tasks of UAP. One is mini-set UAP, which is to train UAP on a mini-set with typically 32 or 64 images. The other is targeted UAP, which is to generate UAP to cheat the classifier to a user-given class.
 
 For mini-set UAP, extra data augmentation techniques like random rotations and random horizontal flips are used to avoid overfitting. The performance is displayed in Table \ref{tab:miniset}. Of course, the FRs of mini-set UAP are lower than that on the full set. But for UAP, the drop is quite small and still FG-UAP is better than the state-of-the-art UAPs under this setting. 
 \begin{table}[H]
\centering
\caption{FRs (\%) of FG-UAP and Cosine-UAP attacking CNNs on mini-sets. $N$ denotes the number of available samples. The best FRs are highlighted in bold. }
\label{tab:miniset}
\vspace{2mm}
\resizebox{\linewidth}{!}{
\begin{tabular}{@{}cccccccc@{}}
\toprule
method     & $N$ & AlexNet & GoogLeNet & VGG16 & VGG19 & ResNet50 & ResNet152 \\ \midrule
Cosine-UAP & 64 & 96.91 & 86.12 & 95.64 & 94.29 & 91.74 & 88.77     \\
FG-UAP     & 64 & \textbf{97.13} & \textbf{88.64} & \textbf{97.55} & \textbf{96.17} & \textbf{94.48} & \textbf{89.24}    \\
Cosine-UAP & 32 & 95.66 & 87.40 & 96.46 & 94.27 & 92.15 & 87.20     \\
FG-UAP     & 32 & \textbf{96.53} & \textbf{87.94} & \textbf{97.20} & \textbf{96.26} & \textbf{93.39} & \textbf{88.63}    \\
 \bottomrule
\end{tabular}}
\end{table}

Original UAPs, including FG-UAPs, belong to untargeted attacks. However, we can also use it for targeted attack, which is promising because 
FG-UAP is based on NC that means the features of different images are similar in the attacked layers and can be led to a specific class. 
To implement targeted attacks, 
we introduce a targeted FG Loss by adding a term in the original FG Loss, i.e., 
\begin{equation}
    \mathcal{L}_{\mathrm{FG}}(\boldsymbol{h}, \boldsymbol{h'}, i) = \mathcal{L}_{\mathrm{FG}}(\boldsymbol{h}, \boldsymbol{h'}) + f'_i,
\end{equation}
where $i$ is the target class, and $f'_i$ denotes the $i$th logit value of the adversarial example. We adopt VGG16 as the victim model, and randomly choose ten classes, along with the dominant class in untargeted FG-UAP, as the target class to craft corresponding FG-UAP. For comparison, we also report the experimental results for state-of-the-art targeted UAP, DF-UAP \cite{dfuap}, with settings claimed in the paper. For targeted UAP, we additionally record the Targeted Fooling Ratio (TFR) \cite{gap}, i.e., only when the output is exactly the targeted class. We vary the targeted class $c_{\mathrm{target}}$ and report the TFRs in Table \ref{tab:target}, while the FRs are also reported for reference. Generally, FG-UAP has consistent superiority on TFRs and the performance of contrast DF-UAP is quite unstable. Another interesting observation is that DF-UAP and FG-UAP both attain outstanding performance on the $109$-th class (brain coral), which is also the class untargeted UAPs aim at, since there is feature gathering and FG-UAP could find this direction in an unsupervised way.

\begin{table}[H]
\centering
\caption{FRs and TFRs of targeted UAPs for VGG16 trained with different UAP methods. }
\label{tab:target}
\vspace{2mm}
\resizebox{\linewidth}{!}{
\begin{tabular}{@{}cc|cccccccccc|c@{}}
\toprule
\multicolumn{2}{c|}{$c_{\mathrm{target}}$}                            & 1     & 100   & 200   & 300   & 400   & 500   & 600   & 700   & 800   & 900   & 109   \\ \midrule
\multicolumn{1}{c|}{\multirow{2}{*}{DF-UAP}}                    & TFR (\%)      & 82.25 & 71.67 & 48.03 & 80.55 & 66.21 & 78.77 & 24.27 & 75.47 & 81.08 & 80.28 & 91.40 \\
\multicolumn{1}{c|}{} & FR (\%) & 93.94 & 91.76 & 90.42 & 91.91 & 88.84 & 91.41 & 92.40 & 89.85 & 92.99 & 93.19 & 97.74 \\
 \midrule
\multicolumn{1}{c|}{\multirow{2}{*}{FG-UAP}}                        & TFR (\%)      & 83.32 & 78.07 & 76.28 & 83.22 & 72.67 & 78.07 & 77.35 & 75.71 & 83.67 & 82.43 & 93.73 \\
\multicolumn{1}{c|}{} & FR (\%) & 95.64 & 94.63 & 94.93 & 94.95 & 94.72 & 93.04 & 93.94 & 94.18 & 95.82 & 94.48 & 98.11 \\
 \bottomrule
\end{tabular}}
\end{table}

\section{Discussion on FG-UAP and NC}\label{sec:property}
The close link between UAP and NC is the motivation of our FG-UAP. 
The good performance shown in the above section actually confirms the link. This section investigates FG-UAP in the view of NC, which can help the understanding of both UAP and NC.

\subsection{Label dominance}\label{sec:label dominance}
When the UAPs attack DNNs, they may lead most of natural images to several specific classes, although not intended as an objective. This phenomenon is called \emph{label dominance}, which is first discovered in \cite{deepfooluap} and further been discussed in \cite{vadillo2022analysis, cosineuap}. Here, we first verify whether the phenomenon also happens for our method. 

To measure the label dominance, we report the percentage of top $k$ most frequently occurred categories account for the predicted labels (denoted as dominance ratio $\mathcal{D}_k$). Furthermore, we also examine whether the predicted class of UAP itself is in the top $k$ categories. It can be observed from Table \ref{tab:dominance} that UAPs for all models attain extremely high dominance ratio (compared with originally 0.1\% for any certain class), and the most frequently occurred category is exactly the corresponding UAP's predicted class. This discovery is consistent with that in \cite{cosineuap}. In the view of NC, it can be concluded that the FG-UAP utilizes the collapse of with-class variability collapse to attack DNNs, and finally results in collapse of between-class variability.


\begin{table}[!htpb]
\centering
\caption{Dominance ratios (\%) of FG-UAP for DNNs. The second row indicates DNN's prediction when feeding FG-UAP as input, and its place of how frequently it occurs in perturbed images' prediction.}
\label{tab:dominance}
\vspace{2mm}
\resizebox{\linewidth}{!}{
\begin{tabular}{@{}cccccccccc@{}}
\toprule
 & ALN & GLN & VGG16 & VGG19 & RN50 & RN152 & DeiT-Ti & DeiT-S & DeiT-B \\ \midrule
 $c_{\boldsymbol{\delta}}$&  721($1^{st}$)   &   109($1^{st}$)   & 109($1^{st}$) & 109($1^{st}$) &  971($1^{st}$)   & 854($1^{st}$) & 815($1^{st}$)  & 828($1^{st}$) &  879($1^{st}$)  \\
$\mathcal{D}_1$& 35.18  &  82.15   & 93.32 & 93.45 &  51.24  &  86.42 & 78.10 & 63.79 & 79.32  \\
$\mathcal{D}_3$& 53.66  &  83.71    & 95.43 & 94.28 &  82.77   & 91.51 & 81.95 & 72.40 &  80.44  \\
$\mathcal{D}_5$& 60.64  &  84.49    & 95.95 & 94.78 &  91.64   & 93.16 &83.29  & 74.12 &  81.09  \\
 \bottomrule
\end{tabular}}
\end{table}

\subsection{Feature collapse of adversarial examples}
 In addition to label dominance, we try to go further into the feature-level to see what is happening. Originally, NC is found for natural images. Since universal perturbations are image-independent, it could be expected that adversarial examples generated by our FG-UAP have more obvious NC. To see that, 
 we calculate the magnitude of the between-class covariance compared with the within-class covariance of the train activations, which is used in \cite{neuralcollapse} as a measure for collapse. Mathematically, we calculate
 $
\operatorname{Tr}(\Sigma_{W} \Sigma_{B}^{\dagger})
$,
 where $\Sigma_{B}$ is the inter-class covariance matrix
 $
\boldsymbol{\Sigma}_{B} \triangleq \underset{c}{\operatorname{Ave}}\left\{\left(\boldsymbol{\mu}_{c}-\boldsymbol{\mu}_{G}\right)\left(\boldsymbol{\mu}_{c}-\boldsymbol{\mu}_{G}\right)^{\top}\right\}
$.

\begin{figure}[H]
    \centering
    \includegraphics[width=\textwidth]{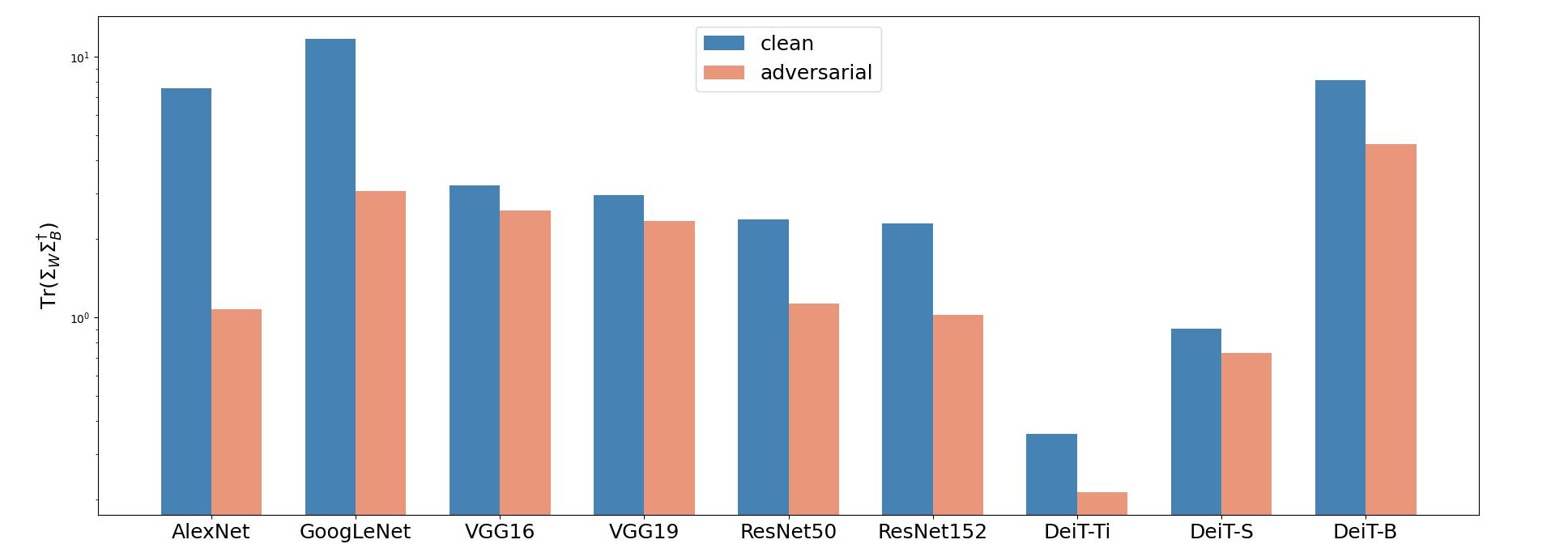}
    \caption{The magnitude of the between-class covariance compared with the within-class covariance of the train activations ($\operatorname{Tr}(\Sigma_{W} \Sigma_{B}^{\dagger})
 $) at the last-layer feature space before and after the images are perturbed.}
    \label{fig:var_ratio}
\end{figure}

The values 
before and after these images are perturbed 
can reflect NC for natural and adversarial examples.
For perturbed images, their predicted classes are used as labels. Results are reported in Figure \ref{fig:var_ratio}, which leads to two main conclusions. 
Firstly, variability collapse indeed happens at the last-layer feature space, since the ratio's value is comparable to that in the original paper. Secondly, the phenomenon of variability collapse gets even worse after these images are corrupted by FG-UAPs. Combined with the label dominance phenomenon in Section \ref{sec:label dominance}, it can be concluded that FG-UAP corrupts images by finding a new direction at the last-layer feature space, which gathers natural images' features to this direction and make the collapse severer. 
As a result, the majority of images are 
predicted as the same class with UAP's.

\subsection{
Redundancy 
in the last-layer space}
According to NC's manifestation, images belonging to the same class gather to a direction at the last-layer space, sharing similar features. This means a small portion of images for one class are quite representative for the majority of images with the same label, from which it follows that fooling them may incidentally fool the others 
as well. 
The redundancy means we only need to consider a smaller scale of images to generate a FG-UAP, not all like regular UAPs do. 

\begin{figure}[H]
    \centering
    \includegraphics[width=.95\textwidth]{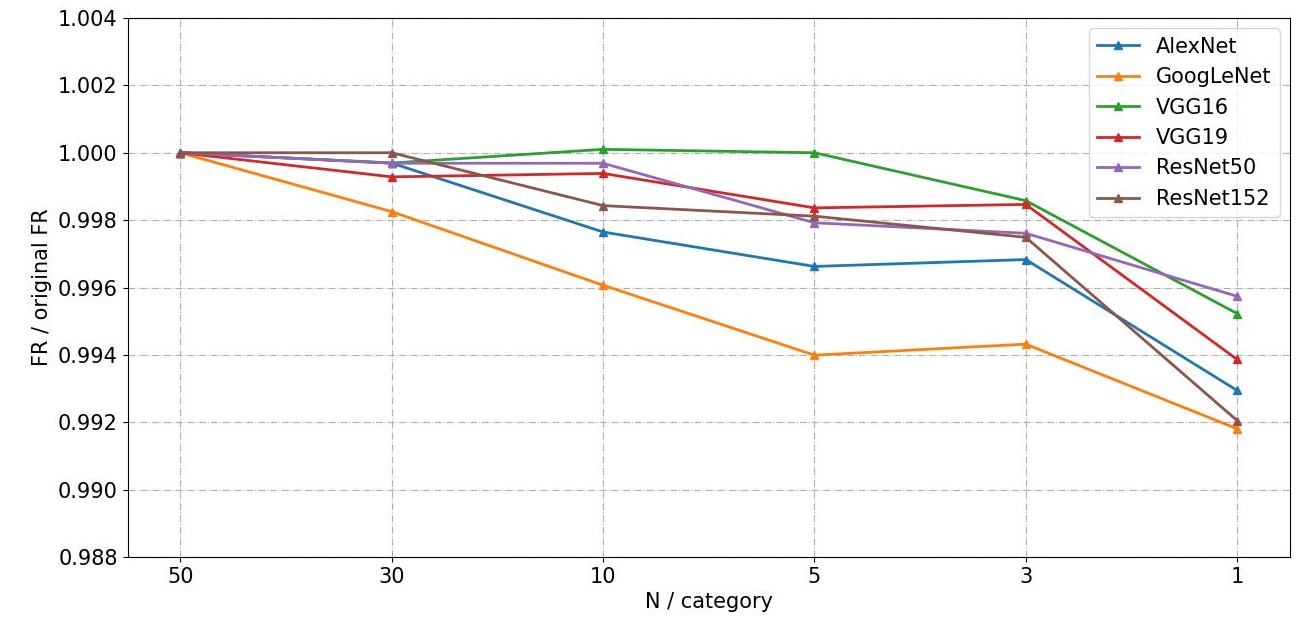}
    \vspace{-5mm}
    \caption{FRs for six victim models when the dataset scale is reduced from fifty to one image per category. The ordinate denotes the ratio of corresponding FR to original FR. }
    \label{fig:FR_decrease}
\end{figure}

Note that here our aim is to maintain comparable performance with limited information for each class, and we do not use extra data augmentation, which is different from section \ref{sec:mini}. We randomly select some samples from each class and train FG-UAP on them, instead of the original 50,000 images. Figure \ref{fig:FR_decrease} displays the extent of FR decrease when the number of selected samples for each class ranges from 50 (original size) to 1. The detailed FR performance of extreme 1,000 images has also been reported in Table \ref{tab:mini}.
Notice that we switch the hyper-parameter to $lr = 0.01 , m = 20$ for better convergence and maintain other settings unchanged. It can be concluded that decreasing the number of images has limited influence on the fooling performance. Even when there is only a single sample for each class, the drop is quite small: the decrease is smaller than 1\% for all the victim models and is still better than the state-of-the-art UAPs, even when they are trained on the full set. This confirms the variability collapse for natural images at the last-layer feature space.

\begin{table}[H]
\centering
\caption{FRs (\%) of FG-UAP attacking CNNs on the full and limited samples. }
\label{tab:mini}
\vspace{2mm}
\resizebox{\linewidth}{!}{
\begin{tabular}{@{}ccccccc@{}}
\toprule
        & AlexNet & GoogLeNet & VGG16 & VGG19 & ResNet50 & ResNet152 \\ \midrule
Full set (50,000)  & 97.77   & 91.53     & 98.45 & 97.77 & 96.23    & 95.59     \\
Mini set (1,000) &  97.08   & 90.78     & 97.98 & 97.17 & 95.82    & 94.83     \\
Deviation      & -0.69   & -0.75  & -0.47 & -0.60 & -0.41    & -0.76     \\ \bottomrule
\end{tabular}}
\end{table}

\section{Conclusion and Future Work}
In this paper, we demonstrate the effectiveness of finding universal attack perturbations at the layer where NC happens. The proposed FG-UAP, which gathers natural images' features with a universal perturbation, 
can achieve 
state-of-the-art performance on the ImageNet dataset 
for both CNNs and ViTs. It also works well in targeted attack or under limited training data. 
Further, we investigate the properties of FG-UAP, which support NC and demonstrate the effectiveness of attacking the features where NC happens. 
The future works include extending our method to other fields, and getting further insight into the relation between existence of UAPs and robustness of DNNs.

\section*{Declaration of Competing Interest}
The authors declare that they have no known competing financial interests or personal relationships that could have appeared to influence the work reported in this paper.

\section*{Acknowledgments}
This work was partially supported by National Natural Science Foundation of China (No. 61977046) and Shanghai Municipal
Science and Technology Major Project (2021SHZDZX0102).

\newpage
\bibliography{reference}

\end{document}